\documentclass{article}

\usepackage[dvipsnames]{xcolor}
\usepackage{microtype}
\usepackage{multirow}
\usepackage{amsfonts}
\usepackage{amsmath}
\usepackage{amssymb}
\usepackage{graphicx}
\usepackage{booktabs} 
\usepackage{subcaption}

\usepackage{tikz}
\usepackage{tkz-graph}
\usetikzlibrary{matrix, positioning, fit, calc}

\usepackage{hyperref}

\usepackage[accepted]{icml2020}

\icmltitlerunning{Neural Bipartite Matching}

\begin{document}

\twocolumn[
\icmltitle{Neural Bipartite Matching}



\icmlsetsymbol{equal}{*}

\begin{icmlauthorlist}
\icmlauthor{Dobrik Georgiev}{cam}
\icmlauthor{Pietro Li\`o}{cam}
\end{icmlauthorlist}

\icmlaffiliation{cam}{
    Department of Computer Science and Technology, University of Cambridge,
    Cambridge, United Kingdom
}

\icmlcorrespondingauthor{Pietro Li\`o}{pietro.lio@cst.cam.ac.uk}
\icmlcorrespondingauthor{Dobrik Georgiev}{dgg30@cam.ac.uk}

\icmlkeywords{}

\vskip 0.3in
]


\printAffiliationsAndNotice{}  

\begin{abstract}
Graph neural networks (GNNs) have found application for learning in the space
of algorithms. However, the algorithms chosen by existing research (sorting,
Breadth-First search, shortest path finding, etc.) usually align perfectly with
a standard GNN architecture. This report describes how neural execution is
applied to a \emph{complex} algorithm, such as finding maximum bipartite
matching by reducing it to a flow problem and using Ford-Fulkerson to find the
maximum flow. This is achieved via neural execution based only on features
generated from a single GNN. The evaluation shows strongly generalising results
with the network achieving \emph{optimal matching} almost 100\% of the time.
\end{abstract}

\section{Introduction}

Many real-world problems can be formulated as graph problems -- social
relations, protein folding, web search, etc. Throughout the years graph
algorithms for solving these tasks have been discovered. One such task is
the problem of finding the maximum flow $f$ from a source to a sink in a graph
$G(V,E)$ whose edges have certain capacities $c(u,v), (u,v)\in E$. (Imagine
material flowing source $\rightsquigarrow$ sink). Any flow must obey two important
properties: the flow on each edge should not exceed the capacity, i.e.
$f(u,v)<c(u,v)$ and for all nodes except source and sink flow should be
preserved, i.e.  $\sum\limits_{v\in V}f(u,v)=\sum\limits_{v\in V}f(v,u)$.
Algorithms for finding maximum flow have found applications in many areas, such
as bipartite matching (attempted here), airline scheduling or image
segmentation \citep{Boykov2006Graph}.

The main topic of this work is evaluating whether graph neural networks (GNNs)
are able to reason like a \emph{complex} algorithm, specifically, whether they
can be used for finding optimal bipartite matching using the Ford-Fulkerson
\citep{FordFulkerson} algorithm for finding maximum flow. Performing the
reasoning is achieved via neural execution, in a similar fashion to
\citet{Velickovic2020Neural}. GNNs have been both empirically
\citep{Velickovic2020Neural} and theoretically \citep{Xu2020What} shown to be
applicable to algorithmic tasks on graphs, \emph{strongly generalising} on
inputs of sizes much larger than trained on. However, these algorithms rely on
a locally contained and fixed dataflow which aligns perfectly with a standard
GNN architecture, making them easy to model with GNNs
\citep[c.f.][]{Xu2020What}.


Our contributions are three-fold: \textbf{1)} We successfully show that GNNs
are suitable for learning a complex algorithm, namely Ford-Fulkerson, which
consists of several \emph{composable} subroutines. To the best of our
knowledge, this is the first time such an algorithm  is neurally executed with
GNNs. \textbf{2)} We demonstrate that GNNs can learn to respect the
invariants of a complex algorithm. \textbf{3)} We devised an evaluation which
not only separately takes into account the accuracy of the subroutines, but
assesses the performance of the Ford-Fulkerson algorithm \emph{as a whole} --
an inconsistency even in one of the subroutines can invalidate the whole
algorithm.

\section{Background}

\subsection{Ford-Fulkerson}

For presentational purposes consider a concise version of Ford-Fulkerson
algorithm given in \citet{CLRS} which operates directly on the residual
graph $G_f$ with residual capacities $c_f$ derived from the input flow graph.
The source and sink of the network are $src$ and $sink$:
\begin{algorithm}[H]
    \caption{Ford-Fulkerson}\label{algo:FF}
    \begin{algorithmic}
        \STATE {\bfseries Input: $G_f$, $src$, $sink$} 
        \WHILE{$\exists$ valid path $p\in G_f$ from $src$ to $sink$}
        \STATE $c_f(p)=\min\{c_f(u,v) : (u,v)\in p\}$
        \FOR {each $(u,v)\in p$}
        \STATE $c_f(u,v)=c_f(u,v) - c_f(p)$
        \STATE $c_f(v,u)=c_f(v,u) + c_f(p)$
        \ENDFOR
        \ENDWHILE
        \STATE {\bfseries return} $\sum\limits_{v\in G} c_f(v, src)$
    \end{algorithmic}
\end{algorithm}

The algorithm above has three key subroutines the neural network has to learn
-- finding augmenting path, finding minimum (bottleneck) capacity on the path
and augmenting the residual capacities along the path.


\subsection{Algorithm Execution} 
\paragraph{Preliminary definitions} The GNN receives
a sequence of $T\in\mathbb{N}$ graphs with the same structure (vertices 
edges), but different features representing the execution of an algorithm.
Let the graph be $G(V,E)$. At each timestep $t\in \{1, ..., T\}$, each node
$i\in V$ has \emph{node features} $\vec{x}_i^{(t)}\in\mathbb{R}^{N_x}$ and each
edge $(i,j)\in E$ has \emph{edge features}
$\vec{e}_{ij}^{(t)}\in\mathbb{R}^{N_e}$. At each step of the algorithm
node-level outputs $\vec{y}_i^{(t)}\in\mathbb{R}^{N_y}$ are produced, which are
later reused in $\vec{x}_i^{(t+1)}$.

\paragraph{Encode-process-decode} The execution of an algorithm proceeds by the
encode process decode paradigm \citep{Hamrick2018Relational}. For each
algorithm $A$, an \emph{encoder network} $f_A$ produces the algorithm-specific
inputs $\vec{z}_i^{(t)}$ The result is then processed using the \emph{processor
network} $P$, which is shared across all algorithms. The processor takes as
input encoded inputs $\mathbf{Z}^{(t)}=\{\vec{z}_i^{(t)}\}_{i\in V}$ and edge
features $\mathbf{E}^{(t)}=\{\vec{e}_{ij}^{(t)}\}_{e\in E}$ to produce latent
features $\mathbf{H}^{(t)}=\{\vec{h}_i^{(t)}\in \mathbb{R}^K\}_{i\in V}$:
Algorithm specific outputs are calculated by its corresponding \emph{decoder
network} $g_A$.  Termination of the algorithm is decided by a \emph{termination
network}, $T_A$, specific for each algorithm. The probability of the
termination of an algorithm is obtained by applying the logistic sigmoid
activation $\sigma$ to the outputs of $T_A$. This is summarised as:
\setlength{\abovedisplayskip}{2pt}
\setlength{\belowdisplayskip}{2pt}
\setlength{\abovedisplayshortskip}{2pt}
\setlength{\belowdisplayshortskip}{2pt}
\begin{align}
    \vec{z}_i^{(t)} &= f_A\left(\vec{x}_i^{(t)}, \vec{h}_i^{(t-1)}\right)\\
    \mathbf{H}^{(t)} &= P\left(\mathbf{Z}^{(t)}, \mathbf{E}^{(t)}\right)\\
    \vec{y}_i^{(t)} &= g_A\left(\vec{z}_i^{(t)}, \vec{h}_i^{(t)}\right)\\
    \tau^{(t)} &= \sigma\left(T_A\left(\overline{\mathbf{H}^{(t)}}\right)\right)
\end{align}
where $\overline{\mathbf{H}^{(t)}}=\tfrac{1}{\lvert V\rvert}\sum_{i\in
V}\vec{h}_i^{(t)}$. The execution of the algorithm proceeds while $\tau^{(t)}
> 0.5$ and $t < \lvert V-1 \rvert$.  The algorithm is \emph{always} terminated
in $\lvert V-1 \rvert$ steps.

\paragraph{Supervising algorithm execution} The aim for every algorithm is to 
learn to replicate the actual execution as close as possible. To achieve this, 
the supervision signal is driven by the actual algorithm outputs at every step
$t$. 

For more details, please refer to
\citet{Velickovic2020Neural}.

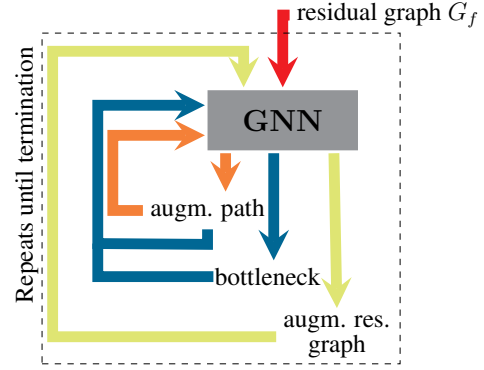
\begin{figure}[t]
    \centering
    \scalebox{0.8}{
        \begin{tikzpicture}
    \tikzset{EdgeStyle/.style = {line width=1.8mm, -stealth, color=Red}}
    \tikzset{EdgeStyleNoArrow/.style = {line width=1.8mm, color=Red}}
    \fill[color=Gray] (-3, -0.5+0.25) -- (-0.5, -0.5+0.25) -- (-0.5, 0.5+0.25) -- (-3, 0.5+0.25) -- (-3, 0.5+0.25);
    \node[align=center] (rn) at (0, 2) {\large residual graph $G_f$};
    \node[align=center] (GNN) at (-1.75, 0.25) {\Large $\mathbf{GNN}$};
    \draw[EdgeStyle] (rn.west) -- (GNN.north |- rn.west) -- ($(GNN.north)+(0,0.25)$);
    \node[align=center] (ap) at (-3, -1.25) {\large augm. path};
    \node[align=center] (bt) at (-2, -2.35) {\large bottleneck};
    \node[align=center] (an) at (-0.85, -3.35) {\large augm. res.\\ \large graph};
    \draw[EdgeStyle, color=Orange] ($(GNN.south -| ap)+(0.3,-0.25)$) -- ($(ap.north)+(0.3,0)$);
    \draw[EdgeStyle, color=Orange] (ap.west) -- ($(ap.west)-(0.5,0)$) -- ($(ap.west |- GNN.west)-(0.5,0.25)$) -- ($(GNN.west)-(0.50,0.25)$);
    \draw[EdgeStyle, color=MidnightBlue] ($(ap.south)$) -- ($(ap.south)-(0,0.25)$) -- ($(ap.west)-(0.75,0.55)$) -- ($(ap.west |- GNN.west)-(0.75,-0.25)$) -- ($(GNN.west)-(0.5,-0.25)$);
    \draw[EdgeStyle, color=MidnightBlue] ($(GNN.south)+(-0.15,-0.25)$) -- ($(bt.north)+(0.1,0)$);
    \draw[EdgeStyleNoArrow, color=MidnightBlue] ($(bt.west)+(0.1,0)$) -- ({$(ap.west)-(0.75,0.55)$}|-bt) -- ($(ap.west)-(0.75,0.55)$);
    \draw[EdgeStyle, color=GreenYellow] ($(GNN.south)+(0.85,-0.25)$) -- (an.north);
    \node (ane) at (an -| {$(ap.west)-(1.5,0.55)$}) {};
    \node (ane2) at (ane |- {$(GNN.north)+(0,0.85)$}) {};
    \node (ane3) at (ane2 -| {$(GNN.north)-(0.25, 0)$}) {};
    \draw[EdgeStyle, color=GreenYellow] (an.west) -- (an -| {$(ap.west)-(1.5,0.55)$}) -- (ane |- {$(GNN.north)+(0,0.85)$}) -- (ane2 -| {$(GNN.north)-(0.65, 0)$}) -- ($(GNN.north)-(0.65,-0.25)$);
    \draw[dashed] (-5.75, -3.8) -- (0.2, -3.8) -- (0.2, 1.70) -- (-5.75, 1.70) -- (-5.75, -3.8);
    \node[rotate=90] (rut) at (-6, -0.65) {\large Repeats until termination};
\end{tikzpicture}
    }
    \vspace{-10pt}
    \caption{
        Neural execution of Ford-Fulkerson: The GNN takes as input a residual
        graph $G_f$. At each step of the algorithm, the GNN computes the augmenting
        path which is then reused (orange) to find the bottleneck edge on
        the path. The bottleneck and the augmenting path are then fed through
        (blue) to produce the residual graph with augmented capacities.
        The resulting residual graph is the input to the next step (yellow).
    }\label{fig:exec_FF}
\end{figure}

\section{Neurally Executing Ford-Fulkerson}

On a high-level, execution proceeds as in Figure \ref{fig:exec_FF}. The neural
network computes an augmenting path from an input residual graph. Then, given
the path, the bottleneck on it is found and the capacities on the path are
changed according to Algorithm \ref{algo:FF}. The resulting new residual graph
is reused as input to the next step and this process repeats until termination
of the algorithm. 

\paragraph{Finding Augmenting Path} One of the key challenges to the task of
finding an augmenting path was deciding how the supervision signal is
generated. Supervising towards algorithms such as Breadth-First/Depth-First
search turned out to bee too difficult to train, since the algorithm and the
learner could choose a different augmenting path (in both cases valid), but the
learner is `penalised' for its decision.

\begin{figure}[h]
    \centering
    \hspace*{-15pt}
    \scalebox{1.15}{\parbox{\linewidth}{%
        \centering
\usetikzlibrary{arrows, shapes, calc, fit, shapes.geometric}
\definecolor{mygreen}{rgb}{0.00,0.5,0.00}
\definecolor{myblue}{rgb}{0,0,0.6}
\definecolor{myred}{rgb}{1,0,0}
\begin{tikzpicture}[scale=0.68]
\tikzset{VertexStyle/.style =
    {draw, shape=circle,minimum size=25pt,inner sep=0pt, scale=0.68}
}
\tikzset{PredStyle/.style = {transform canvas={xshift=1.5mm, yshift=1.5mm}, text=mygreen}}
\tikzset{LabelStyle/.style = {text=black, font=\small, below, scale=0.8, sloped}}
\tikzset{SelectedEdgeStyle/.style = {draw, line width=5pt, mygreen!30}}
\tikzset{EdgeStyle/.style = {->, thick, -stealth}}
\tikzset{LabelStyle/.style = {text=black, font=\small, below, scale=0.67, sloped}}
\begin{scope}[scale=1.0, every node/.style={scale=1.0}, every edge/.style={scale=1.2}]
\node[VertexStyle] (G-res-src) at (-4,0) {src};
\node[VertexStyle] (G-res-sink) at (4,0) {sink};
\foreach \index/\name in {1/1, 2/2}
    \node[VertexStyle] (G-res-\name) at (-1.5,{3*(1.5-\index)}) {$\name$};
\foreach \index/\name in {1/3, 2/4}
    \node[VertexStyle] (G-res-\name) at (1.5,{3*(1.5-\index)}) {$\name$};
\end{scope}
\tikzset{SelectedEdgeStyle/.style = {draw, line width=5pt, myred!30}}
\draw[SelectedEdgeStyle] (G-res-src) to (G-res-1);
\draw[SelectedEdgeStyle] (G-res-1) to (G-res-3);
\draw[SelectedEdgeStyle] (G-res-3) to (G-res-sink);
\tikzset{SelectedEdgeStyle/.style = {draw, line width=5pt, mygreen!30}}
\draw[SelectedEdgeStyle] (G-res-src) to (G-res-2);
\draw[SelectedEdgeStyle] (G-res-2) to (G-res-4);
\draw[SelectedEdgeStyle] (G-res-4) to (G-res-sink);
\draw[EdgeStyle] (G-res-src) to node[LabelStyle]{1/6} (G-res-1);
\draw[EdgeStyle] (G-res-src) to node[LabelStyle]{1/1} (G-res-2);
\draw[EdgeStyle] (G-res-1) to node[LabelStyle]{1/2} (G-res-3);
\draw[EdgeStyle] (G-res-2) to node[LabelStyle]{1/5} (G-res-4);
\draw[EdgeStyle] (G-res-1) to node[LabelStyle]{1/7} (G-res-4);
\draw[EdgeStyle] (G-res-3) to node[LabelStyle]{1/8} (G-res-sink);
\draw[EdgeStyle] (G-res-4) to node[LabelStyle]{1/7} (G-res-sink);

\tikzset{LabelStyle/.style = {text=black, font=\small, above, scale=0.67, sloped}}
\tikzset{EdgeStyle/.style = {<-, thick, stealth-}}
\draw[EdgeStyle, transform canvas={xshift=-1mm, yshift=2mm}] (G-res-src) to node[LabelStyle]{0/12} (G-res-1);
\draw[EdgeStyle, transform canvas={xshift=1mm, yshift=2mm}] (G-res-src) to node[LabelStyle]{0/11} (G-res-2);
\draw[EdgeStyle, transform canvas={yshift=2mm}] (G-res-1) to node[LabelStyle]{0/5} (G-res-3);
\draw[EdgeStyle, transform canvas={yshift=2mm}] (G-res-2) to node[LabelStyle]{0/15} (G-res-4);
\draw[EdgeStyle, transform canvas={xshift=1mm, yshift=1mm}] (G-res-1) to node[LabelStyle]{0/13} (G-res-4);
\draw[EdgeStyle, transform canvas={xshift=1mm, yshift=2mm}] (G-res-3) to node[LabelStyle]{0/1} (G-res-sink);
\draw[EdgeStyle, transform canvas={xshift=-1mm, yshift=1.5mm}] (G-res-4) to node[LabelStyle]{0/1} (G-res-sink);
\end{tikzpicture}%
    }}
    \vspace{-5pt}
\end{figure}
The solution to this problem is presented above. Additional weights are
attached to each edge (edges are in the format \emph{capacity/weight}). Now, if
we choose to find the shortest path\footnote{It is theoretically possible that
two shortest paths exist, but in practice this rarely occurred.}, the bottom
path (green) is preferred over the top one (red). This changes the task from
finding an augmenting path to finding \emph{the shortest} augmenting path,
given the additional weights. Finding the shortest path with the Bellman-Ford
algorithm \citep{b-f} can be achieved by learning to predict predecessors for
each node \citep{Velickovic2020Neural}. The network needs to learn
to ignore zero capacity edges.

\paragraph{Bottleneck Finding} After an augmenting path is found, the next step
is to find the bottleneck capacity along this path. All edges not
on the augmenting path are masked out (deterministically) and each edge is
assigned a probability of being the bottleneck. Inspired by
\citet{Yan2020Neural}, the probabilities were generated using a readout
attention \emph{computed from the messages between edges produced by the GNN
from the last Bellman-Ford timestep}. We have found that a single transformer
encoder layer followed by a fully-connected layer is sufficient for our task.

\paragraph{Augmenting Path Capacities} Assuming integer
capacities\footnote{This does not make the problem less general.} predicting
the edge capacities after the augmentation is achieved using \emph{logistic
regression} over the possible new \emph{forward} capacities. For
each edge $e_i$ with capacity $c_{e_i}$, based on the message generated for
this edge by the GNN, we assign probabilities to each number of the range
$[0;c_{e_i}]$. \emph{Each forward-backward edge capacity pair keeps constant sum.}

To provide unique supervision signal for the above two tasks, random walks of
length 5 are generated, together with random integer edge capacities in the
range [1; 10].

\section{Evaluation through simulation}
\begin{algorithm}[t]
    \caption{Simulated Ford-Fulkerson}\label{algo:simFF}
    {\small
    \begin{algorithmic}
        \STATE {\bfseries Input: $G_f$, $src$, $sink$, $oracle$} \COMMENT{Neural network $oracle$}
        \STATE $cnt_b=1$
        \WHILE{$oracle$.\textsc{find-path}($G_f, src, sink$)} 
        \STATE $p=oracle.path$
        \STATE $c_f(p)=oracle$.\textsc{find-path}($G_f, src, sink, p$) 
        \STATE $real$-$c_f(p)=\min\{c_f(u,v) : (u,v)\in p\}$
        \STATE $cnt_b$++
        \IF{$c_f(p) \neq real$-$c_f(p)$}
        \STATE {\bfseries break}
        \ENDIF
        \IF[$t_b$ to avoid endless loops]{$real$-$c_f(p) = 0$} 
        \IF{$cnt_b > t_b$}
        \STATE {\bfseries break}
        \ENDIF
        \ENDIF
        \STATE $oracle$.\textsc{subtract-bottleneck}($G_f, src, sink, p, c_f(p)$) 
        \STATE $cnt_b=1$
        \ENDWHILE
        \STATE {\bfseries return} $\sum\limits_{v\in G} c_f(v, src)$
    \end{algorithmic}}
\end{algorithm}

Simply evaluating each step separately may not provide sufficient insight on
how well the algorithm is learnt -- discrepancies in either subroutine can
nullify the correctness of the algorithm. Here we present \emph{evaluation
through simulation}, which simulates the Ford-Fulkerson from Algorithm
\ref{algo:FF}. Algorithm \ref{algo:simFF} summarises the simulation.
Subroutine details and design decisions are discussed below.


\paragraph{Finding Augmenting Path and Termination} The main issue with this
step is that it is not possible to distinguish whether a valid path does not
exist or the network is unable to find it. A trivial heuristic is terminating
the algorithm as soon as the network produces an invalid path containing a zero
capacity edge. A slightly better approach is a \emph{thresholding heuristic} --
pre-defining a threshold hyperparameter $t$ and terminating the execution if
the network is unable to find a path $t$ consecutive times. To add some
non-determinism edge weights are randomised for every attempt.

A smarter approach would be to learn to predict which nodes are reachable in
the residual network via edges with positive capacity using the Breadth-First
Search (BFS) algorithm. Therefore we can decide to terminate the algorithm, by
predicting whether the sink is reachable from the source. If predicted
reachable, a possible path from the source to the sink is generated by
predicting predecessors. This heuristic is less artificial than the previous
one, but now we have no guarantee that the generated path is valid. However,
the bottleneck finding subroutine can be used to detect the presence of a zero
capacity edge.

\paragraph{Bottleneck Finding} Similar issue arises here: the network could
predict a wrong edge as the bottleneck on the path, making the algorithm
incorrect. If such an error occurs, the Ford-Fulkerson algorithm is terminated
instantly. Under the bipartite matching setting this can only happen if the
generated path is invalid. In such case if the network correctly predicts
a zero capacity edge, the path-finding step is rerun again. A threshold is used
to avoid endless loops.

\paragraph{Augmenting Path Capacities} The new predicted capacities are
compared against the real ones and if they are different, the Ford-Fulkerson
algorithm is terminated. This may appear as a too strict policy, but evaluation
on the bipartite matching setting showed that the network learns to accurately
perform this step.

\paragraph{Design Motivation} If any of the above subroutines is wrong the flow
value produced will be lower than optimal. Incorrect path-finding will keep
generating invalid paths.  Badly learnt BFS, bottleneck finding or subtraction
can cause premature termination. Additionally, a well-learnt bottleneck finding
will allow for reruns to be generated, allowing the network to `correct'
itself, to some extent.
\begin{table*}[t]%
    \centering
    \caption{
        Accuracy of finding maximum flow at different graph sizes.
        Model format is
        $<$\emph{architecture}$>$($<$\emph{termination-heuristic}$>$).
        Termination heuristic is formatted as ($t=X$), where $X$ is
        pre-determined. PNA-$STD$ denotes PNA without the std aggregator.
    }\label{tab:accs}
    {\small
    \begin{tabular}{l r r r r}
        \toprule
        \multirow{2}{*}{\textbf{Model}} & \multicolumn{4}{c}{\textbf{Accuracy}} \\
        & $1\times$ scale & $2\times$ scale & $4\times$ scale & $8\times$ scale \\
        \midrule
        $\text{\scriptsize MPNN}(t=1)$ & $97\%\pm1.61\%$ & $90\%\pm3.46\%$ & $97.8\%\pm2.44\%$ & $100\%\pm0.00\%$\\
        $\text{\scriptsize MPNN}(t=3)$ & $100\%\pm0.00\%$ & $99.4\%\pm1.28\%$ & $100\%\pm0.00\%$ & $100\%\pm0.00\%$\\
        $\text{\scriptsize MPNN}(t=5)$ & $100\%\pm0.00\%$ & $99.8\%\pm0.6\%$ & $100\%\pm0.00\%$  & $100\%\pm0.00\%$\\
        \midrule
        $\text{\scriptsize MPNN}(\text{\scriptsize BFS})$ & $99.8\%\pm0.6\%$ & $95.6\%\pm2.65\%$ & $98.0\%\pm2.00\%$  & $100\%\pm0.00\%$\\
        $\begin{tabular}{@{}l@{}}\text{\scriptsize PNA\emph{-STD}}(\text{\scriptsize BFS})\end{tabular}$ & $\mathbf{100\%\pm0.00\%}$ & $\mathbf{99.8\%\pm0.6\%}$ & $\mathbf{100\%\pm0.00\%}$  & $\mathbf{100\%\pm0.00\%}$\\
        \bottomrule
    \end{tabular}}
    \vspace{-10pt}
\end{table*}

The code for neural execution and simulation can be found at \url{https://github.com/HekpoMaH/Neural-Bipartite-Matching}.

\section{Evaluation}

\paragraph{Dataset and training details} 300 bipartite graphs are generated for
training and 50 for validation. The probability of generating an edge between
the two subsets was fixed at $p=\tfrac{1}{4}$. Bipartite graph subset size was
fixed at 8 as smaller sizes generated too few training examples. Both subset
were chosen to have the same size, as the maximum flow (maximum matching) is
dictated from the size of the smaller subset. All subroutines are learnt
simultaneously. Adam optimiser \citep{Kingma2015Adam} was used for training
(initial learning rate 0.0005, batch size 32) and early stopping with patience of
10 epochs on the last step predecessor validation accuracy was performed. 
Evaluating the ability to strongly generalise is performed on graphs with
subset size 8, 16, 32 and 64 (50 graphs each). Standard deviations are obtained
over 10 simulation runs.

\paragraph{Architectural details} Two types of GNNs are assessed for their
ability to learn to execute the Ford-Fulkerson algorithm. These are
Message-passing neural networks (MPNN) with maximisation aggregation rule
\citep{Gilmer2017Neural} and Principal Neighbourhood Aggregation (PNA)
\citep{Corso2020Principal} with the standard deviation (std) aggregator
removed\footnote{The std aggregator for tasks with no input noise (such as
algorithms) results in a model which overfits to the data.}. Latent feature dimension was fixed to $K=32$
features. Inputs (capacities and weights) are given as 8-bit binary numbers.
(Infinity is represented as the bit vector 111...1.) Similar to
\citet{Yan2020Neural}, embedding vector $\vec{v}_i$ is learnt for each bit
position. For each $n$-bit input $\vec{x}$, the input feature embedding is
computed as $\vec{\hat{x}}=\sum_{i=0}^{n-1} x_i\vec{v}_i$.

\paragraph{Results and discussion} We report the accuracy of predicting a flow
(matching) equal to the \emph{maximum} one. Table \ref{tab:accs} presents the
accuracy at different scale. Under threshold based execution, only the path
finding is performed neurally, since all generated paths will have edges with
capacity 1.

An exciting observation is that even a threshold of 1, i.e. terminating
Ford-Fulkerson as soon as an invalid path is generated, yields high accuracy --
about 90\% for the $2\times$ scale and more than 95\% for other datasets. In
other words, if a valid path exists, it is likely that the network will find
it. A threshold of 3 gives a noticeable boost in the accuracy and a threshold
of 5 turns out to be sufficient for an almost perfect execution. An MPNN
processor, which uses BFS for termination \emph{and determines the bottleneck
and edge capacities after augmentation} performs better than threshold based
termination when $t=1$ and is slightly worse than other choices of $t$ at
scales $2\times$ and $4\times$. A further ablation study (Appendix
\ref{app:subroutines}) showed that the latter two subroutines have
infinitesimal impact on the accuracy.

\begin{figure}
    \centering
    \includegraphics[width=.95\linewidth]{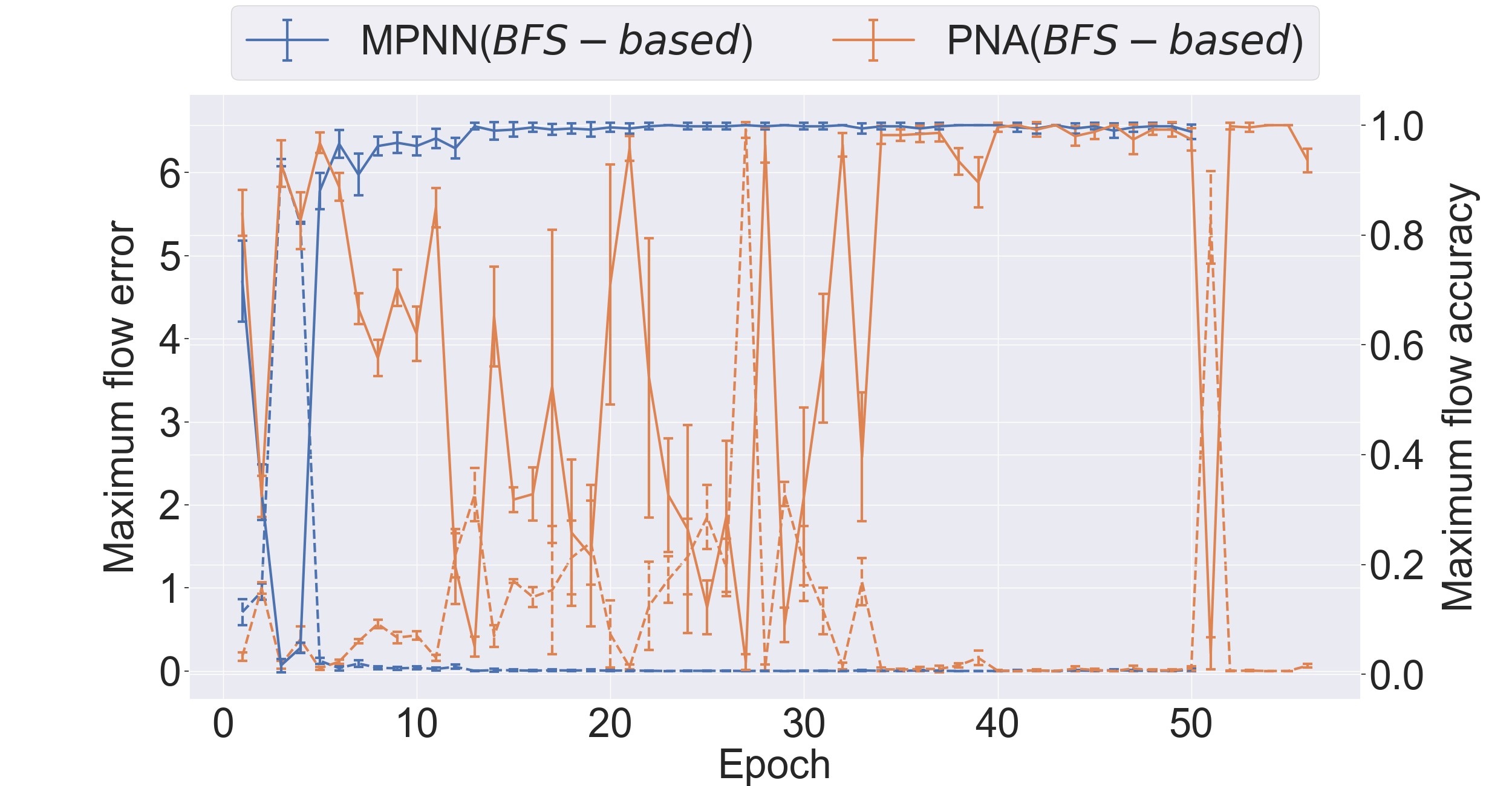}
    \vspace{-10pt}
    \caption{
        Maximum flow accuracy (solid) and mean absolute flow error (dashed) per
        epoch for PNA and MPNN architectures.
    }\label{fig:PNA_acc}
    \vspace{-16pt}
\end{figure}
The best processor architecture was the PNA model, but the std aggregator had
to be removed and the model required extra training data for the BFS task (see
Appendix \ref{app:STDOVERFITS}). PNA also converged much more slowly than MPNN
as can be seen from the flow accuracy per epoch for the 1$\times$ scale, given
in Figure \ref{fig:PNA_acc}. Both networks exhibits some initial instability
during the first epochs, but the MPNN was much more stable (convergence in 10
epochs, compared to 35). Both networks retained near 100\% accuracy once they
converged, with the only exception of epoch 51 for the PNA-based processor.
Both models strongly generalise across all scales.

To further evaluate the strong generalisation ability, the two best-performing
models were tested on bipartite graphs generated with different edge
probability. 50 more test examples were generated for each of \emph{scale}
$\in\{1\times, 2\times\}$ and $p\in\{\frac{1}{5}, \frac{1}{2}, \frac{3}{4}\}$.
Both models performed equivalently and  exhibited average accuracy higher than
99.73\% across all test sets. Further details in Appendix
\ref{app:different_edge_prob}.

We have for the first time shown (near-)perfect strong generalisation for
a complex algorithmic execution task. Based on this, we think that Algorithms
and Deep Learning reinforce each other and we hope this paves to way to further
related applications.

\bibliographystyle{icml2020}

\appendix

\begin{table*}[t]
    \centering
    \caption{
        Accuracy of finding maximum flow at graph sizes. Model format is as in
        Table \ref{tab:accs}.  \emph{-bottle} (minus bottleneck) corresponds to
        changing the neural execution of bottleneck finding to a deterministic
        one.  \emph{-augment} is used when the augmentation step is
        deterministic. \emph{-BFS variety} is for when the PNA-based network is
        trained without the extra data.
    }\label{tab:DONOTREFacc_MPNN}
    {\small
    \begin{tabular}{l r r r r}
        \toprule
        \multirow{2}{*}{\textbf{Model}} & \multicolumn{4}{c}{\textbf{Accuracy}} \\
        & $1\times$ scale & $2\times$ scale & $4\times$ scale & $8\times$ scale \\
        \midrule
        $\text{\scriptsize MPNN}(\text{\scriptsize BFS})$ & $99.8\%\pm0.6\%$ & $95.6\%\pm2.65\%$ & $98.0\%\pm2.00\%$  & $100\%\pm0.00\%$\\
        $\begin{tabular}{@{}l@{}}\text{\scriptsize MPNN}(\text{\scriptsize BFS})\\\emph{-bottle}\end{tabular}$ & $99.8\%\pm0.6\%$ & $96.8\%\pm2.56\%$ & $100\%\pm0.00\%$  & $100\%\pm0.00\%$\\
        $\begin{tabular}{@{}l@{}}\text{\scriptsize MPNN}(\text{\scriptsize BFS})\\\emph{-augment}\end{tabular}$ & $99.8\%\pm0.6\%$ & $97.8\%\pm1.40\%$ & $98.2\%\pm1.66\%$  & $100\%\pm0.00\%$\\
        $\begin{tabular}{@{}l@{}}\text{\scriptsize MPNN}(\text{\scriptsize BFS})\\\emph{-augment}\\\emph{-bottle}\end{tabular}$ & $100\%\pm0.00\%$ & $97.6\%\pm2.33\%$ & $100\%\pm0.00\%$  & $100\%\pm0.00\%$\\
        \midrule
        $\text{\scriptsize PNA}(\text{\scriptsize BFS})$ & $99.4\%\pm0.92\%$ & $50.0\%\pm5.51\%$ & $18.6\%\pm4.73\%$  & $0.2\%\pm0.6\%$\\
        $\begin{tabular}{@{}l@{}}\text{\scriptsize PNA}(\text{\scriptsize BFS})\\\emph{-bottle}\end{tabular}$ & $100\%\pm0.00\%$ & $47.8\%\pm7.67\%$ & $19.4\%\pm3.47\%$  & $0.6\%\pm0.91\%$ \\
        $\begin{tabular}{@{}l@{}}\text{\scriptsize PNA}(\text{\scriptsize BFS})\\\emph{-augment}\end{tabular}$ & $100\%\pm0.00\%$ & $50.2\%\pm5.02\%$ & $18\%\pm3.35\%$  & $0.6\%\pm0.91\%$\\
        $\begin{tabular}{@{}l@{}}\text{\scriptsize PNA}(\text{\scriptsize BFS})\\\emph{-augment}\\\emph{-bottle}\end{tabular}$ & $100\%\pm0.00\%$ & $53.8\%\pm4.69\%$ & $19.2\%\pm4.66\%$ & $0.8\%\pm0.98\%$\\
        \midrule
        $\begin{tabular}{@{}l@{}}\text{\scriptsize PNA\emph{-STD}}(\text{\scriptsize BFS})\\\emph{\scriptsize -BFS variety}\end{tabular}$ & ${100\%\pm0.00\%}$ & ${99.4\%\pm0.92\%}$ & $\color{red!80} 0\%\pm0.00\%$  & $\color{red!80} 0\%\pm0.00\%$\\
        $\begin{tabular}{@{}l@{}}\text{\scriptsize PNA\emph{-STD}}(\text{\scriptsize BFS})\end{tabular}$ & $\mathbf{100\%\pm0.00\%}$ & $\mathbf{99.8\%\pm0.6\%}$ & $\mathbf{100\%\pm0.00\%}$  & $\mathbf{100\%\pm0.00\%}$\\
        \bottomrule
    \end{tabular}}
\end{table*}

\begin{table*}[t]
    \centering
    \caption{
        Accuracy at different edge probability $p$ for the two best models.
    }\label{tab:DONOTREFacc_diff_edge_prob}
    {\small
    \begin{tabular}{c l r r r}
        \toprule
        \multicolumn{1}{c}{\multirow{2}{*}{\textbf{Scale}}} & \multicolumn{1}{c}{\multirow{2}{*}{\textbf{Model}}} & \multicolumn{3}{c}{\textbf{Accuracy}} \\
                                        & & \multicolumn{1}{c}{$p=\frac{1}{5}$} & \multicolumn{1}{c}{$p=\frac{1}{2}$} & \multicolumn{1}{c}{$p=\frac{3}{4}$} \\
        \midrule
        \multirow{2}{*}{$1\times$} & $\text{\scriptsize MPNN}(\text{\scriptsize BFS})$ & $98\%\pm1.55\%$ & $100\%\pm0.00\%$ & $100\%\pm0.00\%$\\
         & $\begin{tabular}{@{}l@{}}\text{\scriptsize PNA\emph{-STD}}(\text{\scriptsize BFS})\end{tabular}$ & $100\%\pm0.00\%$ & $100\%\pm0.00\%$  & $100\%\pm0.00\%$\\
        \midrule
        \multirow{2}{*}{$2\times$} & $\text{\scriptsize MPNN}(\text{\scriptsize BFS})$ & $96.8\%\pm2.03\%$ & $99.4\%\pm0.92\%$  & $100\%\pm0.00\%$\\
         & $\begin{tabular}{@{}l@{}}\text{\scriptsize PNA\emph{-STD}}(\text{\scriptsize BFS})\end{tabular}$ & $99.4\%\pm1.28\%$ & $100\%\pm0.00\%$  & $100\%\pm0.00\%$\\
        \bottomrule
    \end{tabular}}
\end{table*}

\section{Subroutine impact}\label{app:subroutines}

An ablation study of an MPNN based model (Table \ref{tab:DONOTREFacc_MPNN}, top
half) shows that using the network to perform the bottleneck finding and/or
augmentation steps has minimal impact on the overall accuracy: In almost all
cases mean accuracy remains within 1-2\%. This is further supported by the
following two observations. Setting an edge with capacity 0 to be a negative
example and edge with 1 -- positive, the average true negative rate for finding
the bottleneck across all scales is $0.9928$. The average augmentation accuracy
(correctness of capacities after augmentaiton) is $0.9995$. Given these
observations and the fact that the accuracy has a standard deviation of 2.65\%,
the differences could be accredited to the BFS occasionally mispredicting the
sink as unreachable on the last iteration of the Ford-Fulkerson algorithm.

\section{PNA is not highly suitable for the task of graph algorithm execution}\label{app:STDOVERFITS}

\begin{figure}[h]
    \centering
    \includegraphics[width=\linewidth]{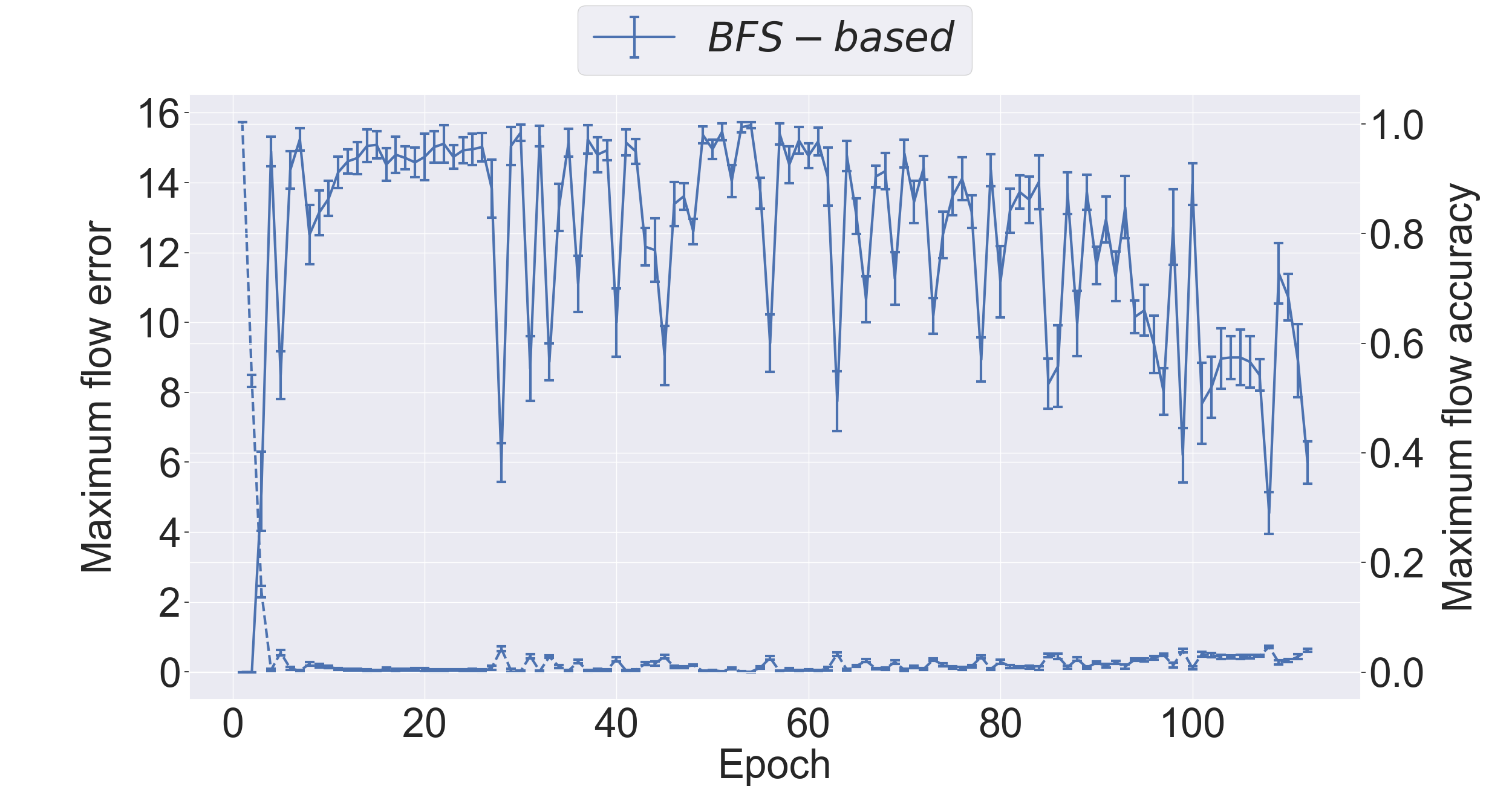}
    \caption{
        PNA on $2\times$ scale. The model shows signs of overfitting.
    }\label{fig:PNASTD_overfits}
\end{figure}

Our initial experiments with the PNA architecture (Table
\ref{tab:DONOTREFacc_MPNN}) did not align with our expectations -- PNA model
performs significantly worse than MPNN on the $2\times$ scale. Plotting the
accuracy per epoch for that scale reveals that the learner initially starts to
converge towards a good solution but it overfits after epoch 25. Our first
hypothesis was that since the task of finding maximum flow is deterministic and
contains no noise, the std aggregator leads to overfitting. Although removing
it did increase the accuracy on the $2\times$ scale, it did not help for strong
generalisation, leading to 0\% accuracy on larger scales.

We already knew that PNA architecture works when BFS is not used. Hence, our
next hypothesis was that since PNA has more parameters and more aggregators
(which do not align to the task) than MPNN, extra training data is needed for
the BFS task. We provided 200 more bipartite graphs drawn from the same
distribution, \emph{but had some pairs (up to 40\%) of nodes matched greedily}.
Although BFS still exhibited some instability, as per Figure \ref{fig:PNA_acc},
it stabilised in the last 10-15 epochs, and produced a model which strongly
generalised.

\section{Varying edge probability}\label{app:different_edge_prob}

Table \ref{tab:DONOTREFacc_diff_edge_prob} shows the accuracy for two best
models on data generated with different edge probability $p$. Higher $p$
produces cases easily solved by both models. The accuracy is less than 100\%
\emph{mainly} for lower edge probability at $2\times$ scale. Both processor
architectures perform equivalently data generated with higher edge probability.

\end{document}